\documentclass[runningheads]{llncs}

\usepackage{amsmath,amssymb,amsfonts}
\usepackage{stmaryrd}
\usepackage{booktabs}
\usepackage{algorithm}
\usepackage{algorithmic}
\usepackage{graphicx}
\usepackage{subfigure} 
\usepackage{float}
\usepackage{color}
\usepackage{comment}
\usepackage{caption}
\usepackage{soul,xcolor}
\usepackage{cite}
\usepackage{multirow}
\usepackage{hyperref}[colorlinks,
            linkcolor=red,
            anchorcolor=blue,
            citecolor=green
            ]
            
\usepackage[symbol]{footmisc}

\newcommand{\tool}{\textsc{NeuroPhysNet}}

\makeatletter

\newcommand{\Rmnum}[1]{\expandafter\@slowromancap\romannumeral #1@}

\makeatother

\usepackage[T1]{fontenc}
\usepackage{graphicx}
\begin{document}
\title{NeuroPhysNet: A FitzHugh-Nagumo-Based Physics-Informed Neural Network Framework for Electroencephalograph (EEG) Analysis and Motor Imagery Classification}
\titlerunning{NeuroPhysNet: PINN for EEG Motor Imagery Classification}
%
\author{Zhenyu Xia\inst{1}\orcidID{0009-0003-7374-7770} \and
Xinlei Huang\inst{1}\orcidID{0000-0002-4761-7652} \and
Yuantong Gu\inst{2}\orcidID{0000-0002-2770-5014}\and
Suvash Saha\inst{1}\orcidID{0000-0002-9962-8919}}
\authorrunning{Z. Xia et al.}
%
\institute{University of Technology Sydney, Sydney, NSW, Australia \\
\email{\{zhenyu.xia, xinlei.huang\}@student.uts.edu.au, suvash.saha@uts.edu.au} 
\and
Queensland University of Technology, Brisbane, QLD, Australia \\
\email{yuantong.gu@qut.edu.au}}
\maketitle              
\begin{abstract}
Electroencephalography (EEG) is extensively employed in medical diagnostics and brain-computer interface (BCI) applications due to its non-invasive nature and high temporal resolution. However, EEG analysis faces significant challenges, including noise, nonstationarity, and inter-subject variability, which hinder its clinical utility.This study introduces {\tool}, a novel Physics-Informed Neural Network (PINN) framework tailored for EEG signal analysis and motor imagery classification in medical contexts. {\tool} incorporates the Fitz Hugh-Nagumo model, embedding neurodynamical principles to constrain predictions and enhance model robustness. Evaluated on the BCIC-IV-2a dataset, the framework achieved superior accuracy and generalization compared to conventional methods, especially in data-limited and cross-subject scenarios, which are common in clinical settings. By effectively integrating biophysical insights with data-driven techniques, {\tool} not only advances BCI applications but also holds significant promise for enhancing the precision and reliability of clinical diagnostics.

\end{abstract}
\section{Introduction}

Electroencephalography (EEG) is a widely used non-invasive technique for measuring the electrical activity of the brain, providing critical insights into various brain functions and disorders. Due to its high temporal resolution and ease of application, EEG has found widespread use in clinical and research settings, such as epilepsy diagnosis, sleep studies, and cognitive monitoring \cite{tatum2018clinical}. One particularly significant application of EEG is in Brain-Computer Interfaces (BCIs), where EEG signals are used to establish direct communication pathways between the brain and external devices. Among BCI paradigms, motor imagery (MI) stands out as a prominent example. MI involves the mental rehearsal of movement without actual physical execution, and it plays a crucial role in developing assistive technologies for individuals with motor impairments \cite{dickstein2007motor}.

Despite its potential in various applications, EEG analysis remains challenging due to the signals’ low spatial resolution, high noise levels, and nonstationarity, which complicate the extraction of meaningful features~\cite{wang2021review}. Traditional neural networks, such as multilayer perceptrons (MLPs) and convolutional neural networks (CNNs), have been widely used to process EEG signals by learning complex patterns from data, often with handcrafted features or transformations like spectral power analysis or spatial covariance matrices~\cite{lu2023lgl}. While effective in some contexts, these approaches face key limitations: they typically lack integration of biophysical priors, reducing interpretability and robustness across datasets~\cite{bijsterbosch2020challenges}; they are prone to overfitting on small medical datasets~\cite{garcia2023machine}; and they struggle to generalize across subjects due to inter-individual variability in brain dynamics~\cite{yu2024learning}.

Physics-Informed Neural Networks offer a promising alternative to address these limitations. The core advantage of PINNs lies in their ability to embed physical laws or domain-specific constraints directly into the learning process, thereby bridging the gap between data-driven methods and mechanistic understanding~\cite{hao2022physics}. PINNs capture EEG’s neurodynamical basis—membrane potential fluctuations and signal propagation—via concise biophysical equations, addressing the physiological constraints often ignored by purely data-driven methods \cite{yang2020neuromorphic}. Embedding these constraints imparts a strong inductive bias that curbs overfitting on limited datasets, enhances robustness to noise and distributional shifts \cite{dou2023machine, hao2022physics}, and affords transparent decision pathways essential for reliable medical applications.

Building on prior PINN advances, we introduce a novel framework for EEG‐based motor imagery classification that embeds the FitzHugh–Nagumo model \cite{izhikevich2006fitzhugh} to enforce neurodynamical constraints and ensure physiologically consistent, interpretable predictions. The primary contributions of this work are as follows:
\begin{itemize}
    \item \textbf{Biophysical Modeling in EEG Analysis:}  
    We introduce a PINN framework that integrates the FitzHugh-Nagumo model, enabling the incorporation of neurodynamical constraints into the analysis of EEG signals. This approach enhances the interpretability and robustness of the model.

    \item \textbf{Advanced Feature Extraction:}  
    A novel feature extraction module is designed to process the biophysically informed outputs of the PINN. This module extracts temporal features at the node level while preserving the physical consistency of the data, resulting in compact and discriminative representations.

    \item \textbf{Application to Motor Imagery Classification:}  
    We demonstrate the effectiveness of the proposed framework in motor imagery classification tasks, achieving improved performance and generalization compared to traditional neural network approaches.
\end{itemize}

\section{Related Work}
\label{sec:related}

Recent advances in deep learning have substantially enhanced the performance of BCIs for MI analysis, particularly through the use of CNNs and Recurrent Neural Networks. Schirrmeister et al. demonstrated the potential of CNNs in EEG decoding and visualization, showcasing the ability of deep learning models to extract meaningful features from EEG signals for improved classification\cite{2017Deep}. Similarly, Lawhern et al. introduced EEGNet, a compact CNN architecture specifically designed for efficient EEG-based BCI applications\cite{2018EEGNet}. Huang et al. also investigated CNN-based deep learning models for MI classification, highlighting their superior performance in classification accuracy\cite{2022EEG}. The low SNR in MI EEG signals poses a challenge in decoding movement intentions, 
but this can be addressed using multi-branch CNN modules that learn spectral-temporal domain features, as suggested in~\cite{jia2023model}. Futhermore, Ju et al. proposed Tensor-CSPNet\cite{2023Tensor} and Graph-CSPNet\cite{ju2022graph}, a novel geometric deep learning framework for motor imagery classification, achieving improved feature extraction, robustness, and interpretability.  

Despite their advances in MI analysis, deep learning approaches remains data-hungry and overfits with scarce labels; its black-box nature hampers interpretability; inter-subject and temporal variability limits generalization; and heavy models slow training and inference, hindering real-time BCI.

To address the limitations of traditional deep learning models, PINNs have emerged as a promising alternative. Raissi et al. first introduced PINNs for solving forward and inverse nonlinear PDEs by embedding physical laws into the learning process; this framework was later adapted to neurodynamics and brain modeling for EEG analysis in BCI applications \cite{2019Physics}. Building on this, Lu et al. developed Physics-Informed DeepONets to handle parametric PDEs and individual variability \cite{wang2021learning}, while Zhang et al. incorporated uncertainty quantification to bolster robustness against noisy EEG data \cite{zhang2019quantifying}. Finally, Jagtap et al. accelerated convergence with adaptive activation functions—crucial for real-time BCI feedback \cite{jagtap2020adaptive}


Our proposed method, the integration of FitzHugh–Nagumo model into a PINN framework, addresses deep‐learning and standard PINN limitations by embedding biophysical constraints that enhance interpretability, reduce reliance on large labeled datasets, improve generalization across subjects, and bolster robustness to EEG noise and individual variability.

\section{Methodology}

In this study, we develop a PINN based on the FitzHugh-Nagumo (FHN) model to analyze and interpret EEG data for BCI applications.

\subsection{FitzHugh-Nagumo Model}

The FHN model, introduced by FitzHugh and Nagumo in the 1960s, offers a two-variable reduction of the Hodgkin–Huxley equations that preserves core action-potential features. By abstracting detailed ionic currents into excitation and recovery variables, it overcomes the high dimensionality and computational burden of the full Hodgkin–Huxley framework, enabling tractable, large-scale simulations and theoretical analyses of neuronal dynamics. Mathematically, the FHN model is articulated through a pair of coupled nonlinear ordinary differential equations that describe the temporal evolution of two critical variables: the activation variable \( u \) and the recovery variable \( v \). These equations are given by:

\begin{equation}
\begin{cases}
\frac{du}{dt} = u - \frac{u^3}{3} - v + I \\
\frac{dv}{dt} = \epsilon (u + a - b v)
\end{cases}
\end{equation}
where $u$ typically represents the membrane potential or the excitatory state of the neuron, while $v$ corresponds to the recovery processes, such as the activation of potassium ion channels that restore the neuron to its resting state after an action potential. $I$ denotes the external stimulus current applied to the neuron, driving its activity. The parameter $\epsilon$ is a small positive constant that signifies the separation of timescales between the fast dynamics of the activation variable and the slower dynamics of the recovery variable. The constants $a$ and $b$ are system parameters that govern the behavior and dynamical characteristics of the model. $t$ represents the time variable, describing the evolution of the system over time, and these differential equations reflect the dynamics of the membrane potential $u$ and the recovery variable $v$ as functions of $t$.

\subsection{Design and Implementation of a Physics-Informed Neural Network Based on the FitzHugh-Nagumo Model}

We propose a three-module PINN architecture—Data Pre-processing, Feature Extraction, and PINN Model—grounded in the FitzHugh–Nagumo equations for EEG analysis.

\subsubsection{Data Preprocessing Module}

The Data Preprocessing Module serves as the initial stage of the PINN architecture, responsible for converting raw EEG time-series data into a structured format that is conducive to subsequent processing and analysis. Specifically, the input EEG data is reorganized into a four-dimensional tensor with the shape \( W \times F \times C \times \omega \), where W represents the number of window slices, F denotes the number of filter banks, C corresponds to the number of EEG channels, and $\omega$ signifies the window length.

This transformation is critical for effectively capturing the spatiotemporal dynamics inherent in EEG signals. The process involves three key steps: temporal segmentation and frequency decomposition.

Temporal segmentation involves dividing the continuous EEG signal into smaller, manageable segments known as window slices. Given an input EEG tensor \( \mathbf{X} \in \mathbb{R}^{B \times C \times T} \), where \( B \) denotes the batch size, \( C \) is the number of EEG channels, and \( T \) represents the total number of time points, the temporal segmentation can be expressed as:

\begin{equation}
\mathbf{X}' = \text{Temporal\_Segmentation}(\mathbf{X}) \in \mathbb{R}^{B \times W \times C \times \omega},
\end{equation}
 where W represents the number of window slices, C corresponds to the number of EEG channels, and $\omega$ signifies the window length. The segmentation aims to divide EEG signals into small segments on the time domain, either with overlapping or without overlapping.

Frequency decomposition is achieved through the application of filter banks, which decompose each windowed EEG signal into multiple frequency passbands. Mathematically, this can be represented as:

\begin{equation}
\mathbf{X}'' = \text{FilterBank\_Decomposition}(\mathbf{X}') \in \mathbb{R}^{B \times W \times F \times C \times \omega},
\end{equation}
where \( F \) denotes the number of filter banks applied. Utilizing causal Chebyshev Type II filters, the raw oscillatory EEG signals are decomposed into distinct frequency bands.

\subsection{Physics-Informed Neural Network Module}

The PINN module is the central component of the proposed architecture, responsible for integrating the FHN model's biophysical constraints with the structured EEG data.

\paragraph{Input Processing and Convolutional Layers}

The model accepts input $\mathbf{X}''$ in the shape \( B \times W \times F \times C \times \omega \).
To align the input data with the convolutional layers, the tensor is reshaped into \( (B \times W) \times F \times C \times \omega \), effectively merging the batch and window dimensions. This reshaped tensor is passed through two 2D convolutional layers, each followed by batch normalization and ReLU activation to stabilize the learning process and introduce non-linearity. The operations are mathematically represented as:
\begin{equation}
\mathbf{X}_1 = \text{ReLU} \left( \text{BatchNorm} \left( \text{Conv2D}(\mathbf{X}'', F_1, k_1, p_1) \right) \right)
\end{equation}
\begin{equation}
\mathbf{X}_1^{\text{pool}} = \text{MaxPool2D}(\mathbf{X}_1, p, s)
\end{equation}
\begin{equation}
\mathbf{X}_2 = \text{ReLU} \left( \text{BatchNorm} \left( \text{Conv2D}(\mathbf{X}_1^{\text{pool}}, F_2, k_2, p_2) \right) \right)
\end{equation}
\begin{equation}
\mathbf{X}_2^{\text{pool}} = \text{MaxPool2D}(\mathbf{X}_2, p, s)
\end{equation}
where \( F_1 \) and \( F_2 \) are the numbers of filters, \( k_1 \) and \( k_2 \) are kernel sizes, and \( p_1 \), \( p_2 \) are paddings.
Pooling layers downsample the feature maps, reducing temporal and spatial dimensions while retaining the most salient features. Here, \( p \) represents the pooling kernel size, determining the dimensions of the pooling operation, and \( s \) represents the stride, which defines the step size for the pooling filter as it slides over the feature map.

\paragraph{Fully Connected Layer}

The flattened output of the convolutional layers is processed by a fully connected layer, which projects the high-dimensional convolutional features into a lower-dimensional hidden representation. Dropout regularization is applied after the fully connected layer to prevent overfitting:
\begin{equation}
\mathbf{h}' = \text{Dropout}\left(\text{ReLU} \left( \mathbf{W}_{\text{fc}} \cdot \text{Flatten}(\mathbf{X}_2^{\text{pool}}) + \mathbf{b}_{\text{fc}} \right)\right),
\end{equation}
where \( W_{fc} \) and \( b_{fc} \) represent the weights and biases of the fully connected layer, respectively. 

\paragraph{Transformer Encoder for Temporal Dependencies}

The hidden representation is passed through a Transformer encoder to model long-range dependencies in the temporal dimension. The Transformer encoder, consisting of multiple layers of multi-head self-attention and feed-forward networks, captures the temporal relationships necessary for decoding EEG signals:
\begin{equation}
\mathbf{H} = \text{TransformerEncoder}(\mathbf{h}')
\end{equation}

The encoder processes the input sequence of embeddings and outputs contextualized representations:
\begin{equation}
\mathbf{H} \in \mathbb{R}^{(B \times W) \times \text{hidden\_dim}}
\end{equation}

\paragraph{Output Layer and Reshaping}

The Transformer output is fed into a linear layer that maps the hidden representation to the activation (\( v \)) and recovery (\( w \)) variables for each neuronal node across time. These variables are fundamental to the FHN model:
\begin{equation}
\mathbf{O} = \mathbf{W}_{\text{out}} \cdot \mathbf{H} + \mathbf{b}_{\text{out}}
\end{equation}

The output tensor $\mathbf{O}$ is reshaped to [batch\_size, sliding\_windows, 2 $\times$ num\_nodes, data\_points], where \( \text{num\_nodes} \) represents the number of nodes in the graph, and \( \text{data\_points} \) denotes the feature dimensions for each node. The first \( \text{num\_nodes} \) along the node dimension is extracted as \( v \), while the remaining \( \text{num\_nodes} \) is extracted as \( w \):
\begin{equation}
\mathbf{v} = \mathbf{O}[:, :, :\text{num\_nodes}, :] \quad \mathbf{w} = \mathbf{O}[:, :, \text{num\_nodes}:, :]
\end{equation}

\paragraph{\textbf{Design Rationale}}

Unlike traditional PINNs based on MLPs, our model employs CNNs and Transformer encoders to exploit EEG’s spatiotemporal structure and accurately estimate the FitzHugh–Nagumo variables $v$ and $w$. Temporal CNNs extract local patterns, stabilized by Batch Normalization and ReLU, while multi-head self-attention captures long-range dependencies to provide global context.

\paragraph{physics-based Loss Function}

Given the predicted values \( v \) and \( w \), the time derivatives are approximated using the finite difference method:
\begin{equation}
\frac{dv}{dt} \approx \frac{v(t + \Delta t) - v(t)}{\Delta t}
\end{equation}
\begin{equation}
\frac{dw}{dt} \approx \frac{w(t + \Delta t) - w(t)}{\Delta t}
\end{equation}
where \( \Delta t \) is the time step.

The residuals for \( v \) and \( w \) are computed as:
\begin{equation}
f_v = \frac{dv}{dt} - \left( v - \frac{v^3}{3} - w + I \right)
\end{equation}
\begin{equation}
f_w = \frac{dw}{dt} - \epsilon (v + a - b w)
\end{equation}

The physics-based loss is then defined as the mean squared error (MSE) of these residuals:
\begin{equation}
\mathcal{L}_{\text{physics}} = \frac{1}{N} \sum_{i=1}^N \left( f_{v,i}^2 + f_{w,i}^2 \right)
\end{equation}
where \( N \) is the total number of data points.

\paragraph{Coupling Term}

To model the interactions among nodes, a coupling matrix \( K \) is introduced. The modified dynamics for \( v \) incorporating coupling are given by:
\begin{equation}
\frac{dv}{dt} = v - \frac{v^3}{3} - w + I + \sum_{j} K_{ij} (v_j - v_i),
\end{equation}
where $K_{ij}$ represents Coupling strength between node \( i \) and node \( j \), $v_j$ is Membrane potential of node \( j \), and $v_i$ is Membrane potential of node \( i \).

The coupling matrix \( K \) is defined as:
\begin{equation}
K_{ij} =
\begin{cases}
\text{coupling\_strength}, & \text{if } i \neq j \\
0, & \text{if } i = j
\end{cases}
\end{equation}
where \( \text{coupling strength} \) is set to \( 0.1 \), and the coupling matrix \( K \), with a shape of \([ \text{num\_nodes}, \text{num\_nodes} ]\), is initialized as an all-ones matrix with its diagonal elements subtracted by 1 and then multiplied by the coupling strength.

The residual for \( v \) including coupling is given by:
\begin{equation}
f_v = \frac{dv}{dt} - \left( v - \frac{v^3}{3} - w + I + \sum_{j} K_{ij} (v_j - v_i) \right)
\end{equation}

\paragraph{Final Physics Loss}
\begin{equation}
\begin{aligned}
\mathcal{L}_{\text{physics}} & = \frac{1}{N} \sum_{i=1}^{N} \Bigg[ 
 \left( \frac{d\hat{v}_i}{dt} - \hat{v}_i + \frac{\hat{v}_i^3}{3} + \hat{w}_i - I \right)^2 \\
& + \left( \frac{d\hat{w}_i}{dt} - \epsilon (\hat{v}_i + a - b \hat{w}_i) \right)^2 
\Bigg],
\label{phyloss}
\end{aligned}
\end{equation}
where \( \hat{v}_i \) and \( \hat{w}_i \) are the predicted membrane potential and recovery variables, respectively. \( I \) denotes the external stimulus, and \( \epsilon \), \( a \), and \( b \) are parameters of the FHN model. For this study, these parameters are set as follows: \( \epsilon = 0.08 \), \( a = 0.7 \), \( b = 0.8 \), and \( I = 0.5 \).

\subsection{Feature Extraction Module}

The Feature Extraction Module processes the outputs of the PINN model, specifically the FHN variables \( v \) (membrane potentials) and \( w \) (recovery variables). The inputs to the module are tensors \( \mathbf{v} \in \mathbb{R}^{B \times N \times T} \) and \( \mathbf{w} \in \mathbb{R}^{B \times N \times T} \), where \( B \) represents the batch size, \( N \) is the number of nodes (e.g., EEG channels), and \( T \) is the number of time points. To align the input tensors with the convolutional layers, the module reshapes and permutes the data into \( \mathbf{v}_{\text{input}}, \mathbf{w}_{\text{input}} \in \mathbb{R}^{(B \cdot N) \times 1 \times T} \). 

The module applies a sequence of one-dimensional convolutional layers to capture localized temporal dependencies. For the input \( \mathbf{v}_{\text{input}} \), the operation of the first convolutional layer is given by:
\begin{equation}
\mathbf{v}_1 = \text{ReLU} \left( \text{BatchNorm} \left( \text{Conv1D}(\mathbf{v}_{\text{input}}, F_1, k, p) \right) \right),
\end{equation}
where \( F_1 \) represents the number of filters, \( k \) is the kernel size, and \( p \) is the padding. To further reduce the temporal resolution, a max-pooling operation is performed:
\begin{equation}
\mathbf{v}_1^{\text{pool}} = \text{MaxPool1D}(\mathbf{v}_1, k_p, s_p),
\end{equation}
where \( k_p \) and \( s_p \) denote the pooling kernel size and stride.

The pooled features are passed through a second convolutional layer with an increased number of filters, refining the temporal features:
\begin{equation}
\mathbf{v}_2 = \text{ReLU} \left( \text{BatchNorm} \left( \text{Conv1D}(\mathbf{v}_1^{\text{pool}}, F_2, k, p) \right) \right),
\end{equation}
followed by another max-pooling operation and the feature maps are flattened into one-dimensional vectors:
\begin{equation}
\mathbf{v}_{\text{flat}} = \text{Flatten}(\text{MaxPool1D}(\mathbf{v}_2, k_p, s_p)),
\end{equation}
and then processed through a fully connected layer to map the extracted temporal features into a latent space:
\begin{equation}
\mathbf{v}_{\text{fc}} = \text{ReLU} \left( \mathbf{W}_{\text{fc}} \cdot \mathbf{v}_{\text{flat}} + \mathbf{b}_{\text{fc}} \right).
\end{equation}

The same operations are applied independently to \( \mathbf{w} \), producing:
\begin{equation}
\mathbf{w}_{\text{fc}} = \text{ReLU} \left( \mathbf{W}_{\text{fc}} \cdot \text{Flatten}(\mathbf{w}_2^{\text{pool}}) + \mathbf{b}_{\text{fc}} \right).
\end{equation}

Finally, the outputs of \( v \) and \( w \) are fused through element-wise addition:
\begin{equation}
\mathbf{f} = \text{LayerNorm}(\mathbf{v}_{\text{fc}} + \mathbf{w}_{\text{fc}}),
\end{equation}
where \(\mathbf{f} \in \mathbb{R}^{B \times N}\) represents the fused feature matrix, which is passed to the next stage of the pipeline.

\section{Experimental Design}
\label{sec:exp}

\subsection{Loss Function and Training} 
The training process was conducted over 100 epochs on a computer equipped with an NVIDIA GeForce RTX 4090 GPU. The total loss is defined as:

\begin{equation}
\mathcal{L} = \mathcal{L}_{\text{classification}} + \lambda \mathcal{L}_{\text{physics}},
\end{equation}
where \( \mathcal{L}_{\text{classification}} \) is the cross-entropy loss, which ensures accurate classification of MI tasks. Meanwhile, \( \mathcal{L}_{\text{physics}} \) enforces the biophysical consistency of the predictions by adhering to the FHN model (see Eq.~\ref{phyloss}). 

During training, a batch size of 64 and an initial learning rate of \( 1e^{-3} \) were used to ensure stable convergence. The weight parameter \( \lambda \) controls the trade-off between the classification and physics-based loss components, allowing the model to balance predictive accuracy with adherence to biophysical principles.

\subsection{Dataset and Baseline Models}
\subsubsection{Dataset}

The BCIC-IV-2a dataset~\cite{tangermann2012review} is a widely used benchmark in EEG-based MI classification research, particularly in the context of BCI development. This dataset contains EEG recordings from nine subjects performing a four-class motor imagery task. The four classes include imagined movements of the left hand, right hand, both feet, and tongue.

\paragraph{Preprocessing}
For the experiments in this study, the EEG signals were preprocessed to enhance the signal-to-noise ratio and extract relevant features. The preprocessing pipeline included band-pass filtering to isolate task-relevant frequency bands (e.g., mu and beta rhythms), artifact removal using EOG channels, and segmentation into fixed-length temporal windows aligned with the onset of motor imagery tasks.

\subsubsection{Baseline Models}

To evaluate the performance and generalization capability of {\tool}, we compared it against a comprehensive set of baseline methods, categorized as follows:

\begin{itemize}
    \item \textbf{CSP-Based Methods:}
    \begin{itemize}
        \item Filter Bank Common Spatial Pattern (FBCSP)~\cite{ang2008filter}: A classical approach that applies spatial filters across multiple frequency bands to extract task-relevant features from EEG signals.
    \end{itemize}

    \item \textbf{Riemannian Geometry-Based Methods:}
    \begin{itemize}
        \item Minimum Distance to Mean (MDM)~\cite{tevet2022human}: Classifies EEG signals by minimizing geodesic distances between covariance matrices.
        \item Temporal Spectral Mapping (TSM)~\cite{lin1811temporal}: Combines temporal and spectral features to improve classification performance.
        \item SPDNet~\cite{huang2017riemannian}: A deep learning model specifically designed to process symmetric positive definite (SPD) matrices.
        \item Tensor-CSPNet~\cite{ju2022tensor}: Extends CSP by incorporating tensor-based feature representations.
    \end{itemize}

    \item \textbf{Deep Learning Architectures:}
    \begin{itemize}
        \item ConvNet~\cite{liu2023novel}: A simple convolutional neural network optimized for EEG feature extraction.
        \item EEGNet~\cite{lawhern2018eegnet}: A compact and efficient architecture designed specifically for brain-computer interface (BCI) applications.
        \item Filter Bank Convolutional Network (FBCNet)~\cite{mane2021fbcnet}: Integrates filter banks with convolutional networks to extract multi-band EEG features.
    \end{itemize}
\end{itemize}

These baselines represent a diverse array of strategies, allowing us to evaluate {\tool}'s performance in comparison to traditional CSP methods, advanced Riemannian geometry-based approaches, and cutting-edge deep learning models.

\begin{table*}[ht]
	\centering
\caption{Comparative Analysis of Subject-Specific Accuracies and Standard Deviations in BCIC-IV-2a Dataset.}
\resizebox{\textwidth}{!}{%
    \begin{tabular}{lccc}

    \toprule
     
        & CV (T) Acc \%  & CV (E) Acc \%  & Holdout (T $\rightarrow$ E) Acc \%  \cr
        
    \cmidrule(lr){1-4}
		FBCSP~\cite{ang2008filter}         & 71.29  & 73.39  & 66.13  \cr
    \cmidrule(lr){1-4}
		EEGNet~\cite{lawhern2018eegnet}    & 69.26  & 66.93  & 60.31  \cr
		ConvNet~\cite{liu2023novel}        & 70.42  & 65.89  & 57.61  \cr
        FBCNet~\cite{mane2021fbcnet}           & 75.48  & 77.16  & 71.53  \cr
    \cmidrule(lr){1-4}
		MDM~\cite{tevet2022human}          & 62.96  & 59.49  & 50.74  \cr
		TSM~\cite{lin1811temporal}         & 68.71  & 63.32  & 49.72  \cr
        SPDNet~\cite{huang2017riemannian}      & 65.91  & 61.16  & 55.67   \cr
		Tensor-CSPNet~\cite{ju2022tensor}  & 75.11  & 77.36  & 73.61  \cr
        ${\tool}^{(20, 4)}$                    & \textbf{76.23}  & \textbf{78.03}  & \textbf{74.20}  \cr
    \bottomrule
    
    \end{tabular}
}
\label{tab:SOTA Comparison}
\end{table*}

\section{Experimental Results}
\label{sec:result}

\subsection{{\tool} Generalization Performance vs. Leading-Edge Methodologies} 

To evaluate the performance and generalization capability of NeuroPhysNet, we compared it against a range of baseline models across cross-validation (CV) and Holdout scenarios on the BCIC-IV-2a dataset. Table~\ref{tab:SOTA Comparison} summarizes the results, which include accuracy metrics from the Training Session (T) and Evaluation Session (E).

Filter Bank Common Spatial Pattern (FBCSP)~\cite{ang2008filter}, a classical CSP-based method, achieves CV accuracies of 71.29\% (T) and 73.39\% (E), and a Holdout accuracy of 66.13\%. While its ability to extract task-relevant spatial and frequency features is notable, its performance is surpassed by most modern methods due to its inability to model complex temporal and physiological dynamics.

For deep learning-based approaches, ConvNet~\cite{liu2023novel} achieves CV accuracies of 70.42\% (T) and 65.89\% (E), and a Holdout accuracy of 57.61\%. EEGNet~\cite{lawhern2018eegnet} performs slightly worse, with CV accuracies of 69.26\% (T) and 66.93\% (E), and a Holdout accuracy of 60.31\%. FBCNet~\cite{mane2021fbcnet}, leveraging filter banks to extract multi-band features, achieves CV accuracies of 75.48\% (T) and 77.16\% (E), and a Holdout accuracy of 71.53\%. Despite its superior frequency representation, it does not incorporate physiological dynamics, which constrains its generalization capabilities.

Among Riemannian geometry-based methods, Minimum Distance to Mean (MDM)~\cite{tevet2022human} achieves CV accuracies of 62.96\% (T) and 59.49\% (E), with a Holdout accuracy of 50.74\%. Similarly, Temporal Spectral Mapping (TSM)~\cite{lotte2018review} records CV accuracies of 68.71\% (T) and 63.32\% (E), and a Holdout accuracy of 49.72\%. SPDNet~\cite{huang2017riemannian} shows slightly better performance than MDM, with CV accuracies of 65.91\% (T) and 61.16\% (E), and a Holdout accuracy of 55.67\%. However, these methods struggle to handle the high-dimensional temporal dynamics inherent in EEG signals, resulting in comparatively low accuracy across all scenarios.

Tensor-CSPNet~\cite{ju2022tensor}, which extends CSP with tensor-based feature representations, achieves CV accuracies of 75.11\% (T) and 77.36\% (E), and a Holdout accuracy of 73.61\%. 

NeuroPhysNet, on the other hand, achieves the highest performance among all methods in multiple scenarios. It records CV accuracies of 76.23\% (T) and 78.03\% (E), and a Holdout accuracy of 74.20\%.

\section{Evaluating the Robustness of NeuroPhysNet with Full and Limited Training Data}

To evaluate the robustness and performance of NeuroPhysNet under varying training data proportions, we conducted experiments on the BCIC-IV-2a dataset. These experiments compared NeuroPhysNet and Tensor-CSPNet using the full training dataset (100\%) and subsets comprising 80\%, 50\%, and 30\% of the training data.

\begin{table*}[h]
    \centering
    \caption{Performance Comparison of NeuroPhysNet and Tensor-CSPNet on BCIC-IV-2a Dataset with Varying Training Data Proportions}
    \resizebox{\textwidth}{!}{%
    \label{tab:comparison_results}
    \begin{tabular}{ccccc}
    \toprule
        \textbf{Data Proportion} & \textbf{Model}         & \textbf{CV Accuracy (T)} & \textbf{CV Accuracy (E)} & \textbf{Holdout Accuracy}  \cr 
        \cmidrule(lr){1-5}
        \multirow{2}{*}{100\% } 
                                 & {\tool}           & \textbf{75.85\%}         & \textbf{76.23\%}         & \textbf{74.20\%}            \cr 
                                 & Tensor-CSPNet          & 75.11\%                  & 77.36\%                  & 73.61\%                           \cr 
        \multirow{2}{*}{80\%}   
                                 & {\tool}           & \textbf{74.12\%}         & \textbf{74.89\%}         & \textbf{72.15\%}             \cr 
                                 & Tensor-CSPNet          & 73.45\%                  & 75.64\%                  & 71.32\%                               \cr 
        \multirow{2}{*}{50\%}   
                                 & {\tool}           & \textbf{65.23\%}         & \textbf{66.72\%}         & \textbf{64.81\%}             \cr 
                                 & Tensor-CSPNet          & 62.45\%                  & 64.31\%                  & 59.72\%                             \cr 
        \multirow{2}{*}{30\%}   
                                 & {\tool}           & \textbf{59.41\%}         & \textbf{60.89\%}         & \textbf{58.45\%}             \cr 
                                 & Tensor-CSPNet          & 56.12\%                  & 58.43\%                  & 53.01\%                              \cr 

    \bottomrule
    \end{tabular}
    }
\end{table*}

The results in Table~\ref{tab:comparison_results} reveal a significant performance advantage for NeuroPhysNet over Tensor-CSPNet across all training data proportions, particularly when the data is limited. When using the full training dataset, NeuroPhysNet achieves a CV accuracy of 76.23\% (E) and 75.85\% (T) and a Holdout accuracy of 74.20\%, slightly outperforming Tensor-CSPNet’s 77.36\% (E), 75.11\% (T), and 73.61\% (Holdout). As the training data proportion decreases, the gap between the two models widens considerably.

At 50\% training data, NeuroPhysNet achieves a CV accuracy of 66.72\% (E) and 65.23\% (T) and a Holdout accuracy of 64.81\%, while Tensor-CSPNet falls to 64.31\% (E), 62.45\% (T), and 59.72\% (Holdout). This substantial drop in Tensor-CSPNet's performance demonstrates its higher sensitivity to data scarcity. In contrast, NeuroPhysNet's integration of FHN equations helps it maintain better generalization by leveraging the physical constraints encoded within its architecture.

At 30\% training data, the differences become even more pronounced. NeuroPhysNet achieves a CV accuracy of 60.89\% (E), 59.41\% (T), and a Holdout accuracy of 58.45\%, outperforming Tensor-CSPNet's 58.43\% (E), 56.12\% (T), and 53.01\% (Holdout) by significant margins. Furthermore, Tensor-CSPNet exhibits less stable predictions across different subjects, which reflects its higher variability compared to NeuroPhysNet.

\subsection{Evaluating the Impact of $v$ and $w$ Features in NeuroPhysNet}

\begin{figure}[ht]
    \centering
    \includegraphics[width=0.6\textwidth]{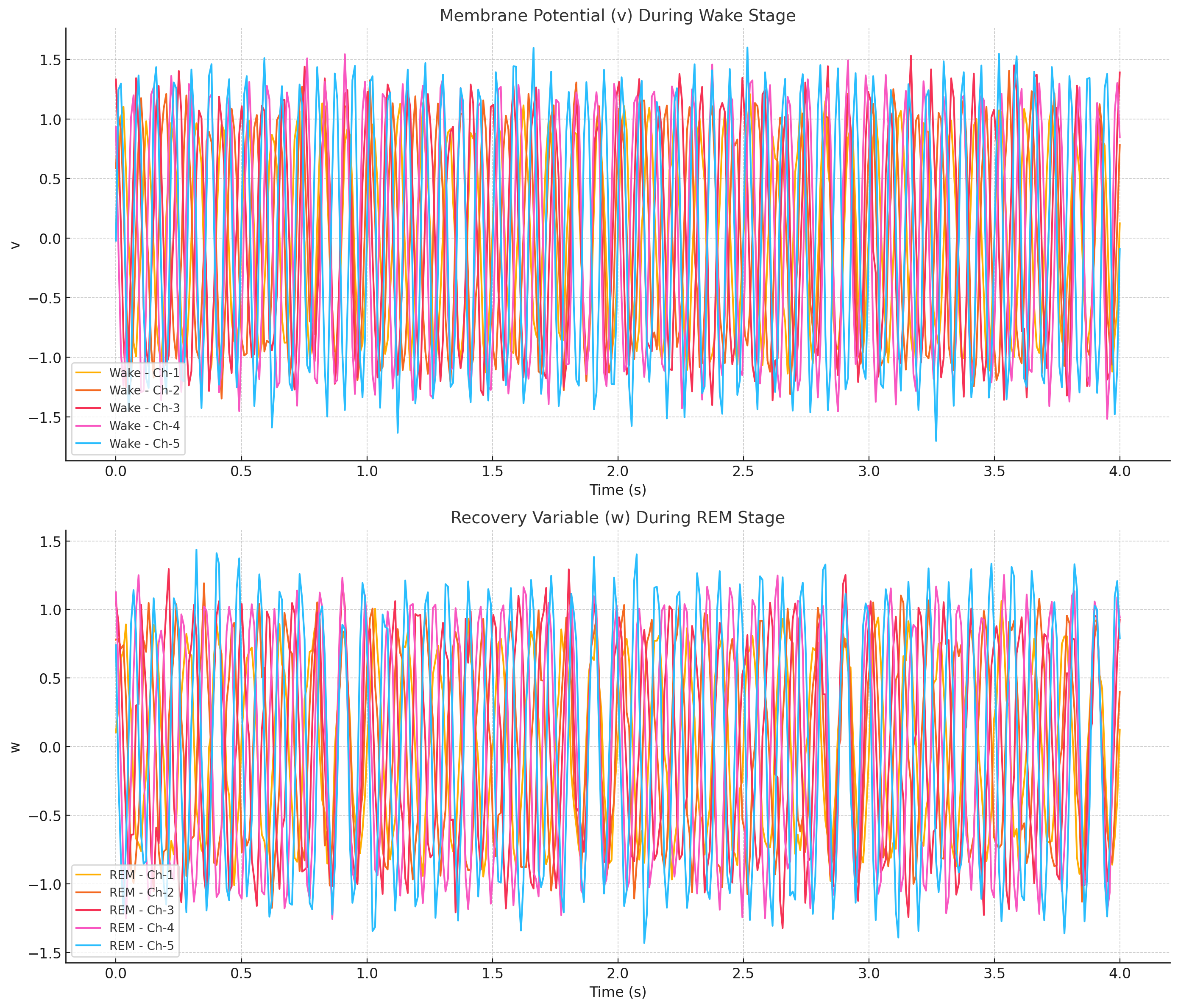}
    \caption{Visualization of $v$ (membrane potential) during the Wake stage and $w$ (recovery variable) during the REM stage for selected EEG channels. }
    \label{fig:v_w_visualization}
\end{figure}

To assess the individual contribution of the physiological features computed by the FHN equations, an experiment was conducted by training and testing NeuroPhysNet using only the membrane potential ($v$) and recovery variable ($w$) as input features, as shown in Figure~\ref{fig:v_w_visualization}. 

\begin{table*}[!h]
    \centering
    \caption{Performance of NeuroPhysNet Using Only $v$ and $w$ Features on BCIC-IV-2a Dataset}
    \label{tab:vw_results}
    \resizebox{\textwidth}{!}{%
    \begin{tabular}{cccc}
    \toprule
        \textbf{Data Proportion} & \textbf{CV Accuracy (T)} & \textbf{CV Accuracy (E)} & \textbf{Holdout Accuracy} \\ 
        \cmidrule(lr){1-4}
        100\%  & 60.34\% & 61.23\% & 59.87\% \\ 
        80\%         & 58.12\% & 58.94\% & 57.43\% \\ 
        50\%         & 54.98\% & 54.32\% & 52.89\% \\ 
        30\%         & 50.11\% & 50.89\% & 49.12\% \\ 
    \bottomrule
    \end{tabular}
    }
\end{table*}

The results presented in Table~\ref{tab:vw_results} show that using only $v$ and $w$ features, NeuroPhysNet achieves reasonable classification performance, though it is significantly lower than the full-feature model. With the full training dataset, the model achieves a CV accuracy of 61.23\% (E) and a Holdout accuracy of 59.87\%. As the training data proportion decreases, performance declines but remains relatively stable. For example, at 30\% training data, the CV accuracy is 50.89\% (E) and 50.11\% (T), and the Holdout accuracy is 49.12\%.
The results indicate that $v$ and $w$ features, derived from the FitzHugh-Nagumo equations, provide valuable discriminative information for EEG classification. Despite the reduced accuracy compared to the full-feature model, the performance remains relatively consistent across different data proportions.

\section{Conclusion}

In this study, we evaluated NeuroPhysNet—a PINN‐based EEG classifier that integrates FHN dynamics—by examining the standalone and combined contributions of the membrane potential $v$ and recovery variable $w$. These physiological features proved robust in data‐limited scenarios and, when fused with conventional EEG features, yielded significant accuracy gains. Compared to previous networks, NeuroPhysNet demonstrated superior accuracy and generalization. Visualization of $v$ and $w$ across sleep stages further confirmed their discriminative power. Our results highlight the promise of PINNs for EEG classification under constrained data conditions and point toward future work on refined fusion strategies, alternative physiological models, and broader EEG applications.

\begin{credits}

\end{credits}
%
%
%
 \bibliographystyle{splncs04}
 \bibliography{ref}

\end{document}